\documentclass[10pt]{article} 
\usepackage[preprint]{tmlr}

\usepackage{amsmath,amsfonts,bm}

\def\eqref#1{equation~\ref{#1}}

\def\1{\bm{1}}

\DeclareMathAlphabet{\mathsfit}{\encodingdefault}{\sfdefault}{m}{sl}
\SetMathAlphabet{\mathsfit}{bold}{\encodingdefault}{\sfdefault}{bx}{n}

\newcommand{\E}{\mathbb{E}}

\DeclareMathOperator*{\argmin}{arg\,min}

\usepackage{hyperref}
\usepackage{url}
\usepackage{amsmath}
\usepackage{amssymb}
\usepackage{multirow}
\usepackage{algorithm}
\usepackage{algpseudocode}
\usepackage{sty/notations}

\usepackage{xcolor}

\usepackage[colorinlistoftodos,prependcaption,textsize=tiny]{todonotes}

\title{Deep End-to-End Survival Analysis with \\ Temporal Consistency}

\author{
\name Mariana Vargas Vieyra\thanks{Work done while at EPFL.}
\email mariana.vargas@rhizome-labs.com \\
\addr Rhizome Labs\\
\AND
\name Pascal Frossard
\email pascal.frossard@epfl.ch \\
\addr École Polytechnique Fédérale de Lausanne
}

\def\month{MM}  
\def\year{YYYY} 
\def\openreview{\url{https://openreview.net/forum?id=XXXX}} 

\begin{document}

\maketitle

\begin{abstract}
    In this study, we present a novel Survival Analysis algorithm designed to efficiently handle large-scale longitudinal data. Our approach draws inspiration from Reinforcement Learning principles, particularly the Deep Q-Network paradigm, extending Temporal Learning concepts to Survival Regression. A central argument in our method relies on temporal consistency, a recursive consistency condition between survival predictions made at consecutive states.
Our framework introduces a delayed target network that decouples the generation of temporally consistent training targets from model updates, providing a more stable training signal while enabling mini-batch optimization and end-to-end training. 
Additionally, it supports arbitrarily complex architectures, enabling the modeling of intricate temporal dependencies.
Through numerous experiments we provide empirical evidence demonstrating our framework’s ability to exploit temporal consistency across datasets of varying sizes. 
Moreover, our algorithm outperforms benchmarks on datasets with long sequences, demonstrating its ability to capture long-term patterns. Finally, ablation studies show that the degree of target delay has a substantial effect on predictive performance and training variability. Across two distinct survival model families, delayed target updates consistently improve over the same-timescale baseline.
\end{abstract}


\section{Introduction}
\label{sec:introduction}

Survival Analysis (SA)~\cite{kalbfleisch2002statistical} is a statistical framework concerned with the study of the time span from when a subject enrolls in a study to the time an event of interest occurs. 
Even though it has been widely used in the medical community~\cite{lee1997survival} it can have numerous applications in other domains such as churn prediction~\cite{Kvamme2019TimetoEventPW,ren2019deep}. 

In the context of Survival Analysis we are presented with a dataset of observations, each of which is associated with a \emph{survival time}.
This type of data is referred to as \emph{time-to-event} data.
In the most common setting, the practitioner is tasked with regressing the life span of each subject within a fixed time window. 
It is not uncommon that, for certain subjects, no event is observed within the study period.
These data points are called \emph{right-censored}, or \emph{censored} for short, and represent incomplete information about the event of interest.
In the dynamic Survival Analysis setting we are presented with time series data rather than vector data.
A common example is clinical trials, where daily health indicators are collected for patients.
This longitudinal setting is arguably richer than the static one because we have information about the evolution of subjects through time. 
In this context, estimations of the survival status of the subjects are updated as new information becomes available, rather than relying on a snapshot of their condition.

Recently, a new approach to estimate survival models in dynamic settings has been proposed by~\cite{TemporallyConsistentSA}. 
Their algorithm, named Temporally Consistent Survival Regression (TCSR), introduces a discrete-time model that builds upon the intuition that longitudinal trajectories can be considered sequences of states drawn from a Markov chain, 
with the event of interest represented as an absorbing terminal state.
The authors then leverage ideas from temporal-difference learning \cite{sutton1988learning} to propose a method that hinges on the idea of temporal consistency. 
That is, the predicted hazard in some specific state should agree, on average, with the shifted hazard predictions in the successor state.
TCSR relies on a nested optimization scheme consisting of an inner loop and an outer loop.
At each training step, the inner iteration queries the current model to obtain a set of "soft targets", and
the outer iteration solves a new survival regression problem against these targets until convergence.
Despite the proven success of TCSR in small scale survival problems, this double-loop procedure can introduce a computational bottleneck when scaling to larger datasets or complex architectures requiring end-to-end training.
\citet{TemporallyConsistentSA} also introduce a variant named "Incremental TCSR" (Inc-TCSR) that partially addresses TCSR's limitations. 
Inc-TCSR collapses the double-loop procedure into a single loop where, at each iteration, the same rapidly changing model is used both to generate the soft targets and to fit them.
Even though this extension enables mini-batch processing, coupling the model's bootstrapping of the training signal and its fitting is a known source of training instability \citep{mnih2015humanlevel}.
These two different mechanisms create a tension between TCSR and its incremental variant. 
The former clearly separates the soft target construction from the model fitting, at the expense of being cumbersome to apply to complex architectures or mini-batch processing.
The latter allows end-to-end optimization by collapsing these two time scales, but this turns the training signal into a moving target, potentially introducing instability.

The main goal of this work is to close the gap between scaling temporally consistent survival analysis to arbitrary deep learning architectures and datasets, and providing a stable training signal.
More specifically, we address the question of whether we can restore a separation between bootstrapping soft targets and updating the models' weights, while retaining the ability to do end-to-end optimization.
Drawing inspiration from Deep Reinforcement Learning \cite{mnih2013playing,lillicrap2016continuous,mnih2015humanlevel} we present a novel framework that introduces a \emph{target network} that mirrors the structure of the model's parameterizing network. 
The aim of the target network is to generate stable target signals for supervising the model's main network, providing more control over the frequency of target updates\citep{fellows2023target}. 
By keeping the targets fixed for longer periods, we not only stabilize the learning process but also enable efficient handling of data in batches. 
Furthermore, the model's parameters can be fitted end-to-end, without requiring re-training at every target update. 
This capability permits to accommodate more expressive architectures that can better model intricate temporal patterns and long-term dependencies.
Notably, our method can be easily incorporated into parametric survival models that predict a hazard function or a conditional event-time distribution.
We update the model's parameters using Stochastic Gradient Descent (SGD),
and the target network using a slow moving average of the main network.
Our experiments highlight the benefits of decoupling the model updates from the construction of the bootstrapped training signal. 
In particular, we show that introducing a slower-moving target model improves survival estimation performance and reduces the variability of estimates compared to updating bootstrap targets on the same time scale as the online model. These results suggest that separating the two update time scales can facilitate temporally consistent survival learning with stochastic, end-to-end optimization.

Our contributions can be summarized as follows: 
(i) We introduce a novel framework that closes the gap between training stability and scalability in temporally consistent survival analysis.
Our method applies most directly to parametric/distributional survival models that expose a discrete hazard or event-time distribution and can be trained with probabilistic soft targets.
(ii) By leveraging a target network that acts as a delayed copy of the main model's parameters, we are able to bootstrap "slow moving targets." 
This helps stabilize training and facilitates batch processing of the data while maintaining high performance.
(iii) Our algorithm can be trained in an end-to-end fashion and seamlessly incorporates expressive neural network architectures that can better capture complex temporal relationships in large sequences. 
(iv) Finally, we empirically show that our framework can effectively build on temporal consistency in datasets of larger sizes, outperforming classical baselines\footnote{Code: https://github.com/marianaw/deeptcsr}.


\section{Related work}
\label{sec:related_work}

\paragraph{Survival models for longitudinal data.}
Dynamic Survival Analysis is concerned with processing longitudinal data for survival prediction.
These methods generally include classical landmarking \citep{anderson1983analysis,landmarking} and joint modeling approaches that leverage deep learning architectures.
In this work we focus on discrete-time models, where models directly parameterize the conditional hazard function at a set of time points.
\cite{gensheimer2019scalable} proposed such a formulation under the name Nnet-survival, while Kvamme and Borgan \citep{kvamme2021continuous} later studied this class systematically and referred to the hazard-based variant as Logistic-Hazard.

Early approaches for longitudinal time-to-event datasets leverage Recurrent Neural Networks (RNNs) for processing sequential data.
\cite{pmlr-v146-nagpal21a} builds upon the Deep Survival Machine \citep{nagpal2021deep} by adding time-dependant feature representations learned with an RNN while
\cite{lee2019dynamic} introduce Dynamic Deep Hit, a dynamic survival algorithm for competing risks that combines a recurrent architecture with cause-specific networks.
More recent work explores alternative representations of longitudinal trajectories.
\cite{moon2022survlatent} leverages Ordinary
Differential Equation-based Recurrent Neural Networks to compute embeddings of trajectories.
\cite{lin2022deep} leverage a transformer architecture to jointly model sequences and their survival status.
DySurv uses conditional variational inference to learn latent representations of longitudinal data for dynamic survival prediction\citep{mesinovic2023dysurv}, whereas the Longitudinal Transformer for Survival Analysis (LTSA) algorithm employs causal Transformer-based sequence modeling to predict discrete hazard distributions from irregularly sampled longitudinal imaging data \citep{Holste2024_LTSA}.
Taking a different perspective, \cite{kim2023survival} leverages Permanental Processes to propose a non-parametric estimator of the hazard function. 
These approaches primarily differ in how the longitudinal history and event-time distribution are modeled, and do not explicitly use the successor-state temporal-consistency relation as a source of bootstrapped supervision.

\paragraph{Temporal-difference learning for survival analysis.}

\cite{TemporallyConsistentSA} draw inspiration from Temporal Difference Learning (TD Learning) \citep{sutton1988learning} and introduce Temporally Consistent Survival Regression (TCSR).
This algorithm combines the observed survival outcomes with predictions bootstrapped from successor states, leveraging the consistency relation between survival predictions at consecutive states.
TCSR consists of a double iteration process where the inner iteration constructs pseudo-targets, used to refit a survival model.
This procedure is repeated until convergence.
Because at each outer iteration the pseudo-targets remain fixed until the survival model converges, this is a \emph{fitted} TCSR algorithm.
The authors also present Incremental TCSR, an extension to TCSR that bootstraps soft targets and updates the model's weights within the same stochastic optimization loop.
While fitted TCSR separates target construction from model fitting, Inc-TCSR couples the model updates with a continuously changing bootstrap target, a setting closely related to the moving-target problem studied in Temporal Difference learning with function approximation.
A prominent approach to mitigate this problem in Deep Reinforcement Learning is the use of a target network, introduced in DQN by \cite{mnih2015humanlevel} and later extended to slowly updated target networks by \cite{lillicrap2016continuous}.
More recently, \cite{fellows2023target} formalize target-network methods as partially fitted Temporal Difference algorithms and analyze their stabilizing effect, while \cite{che2024target} establish convergence results for target-network TD under function approximation in specific settings.
Our approach builds on this idea by introducing a slowly updated target model into incremental temporally consistent survival regression.



\section{Preliminaries}
\label{sec:preliminaries}

In this Section we introduce foundational concepts that are relevant to our method. More specifically, we discuss dynamic Survival Analysis, its formulation as a classification problem, and how temporal consistency can improve survival models in terms of robustness and accuracy.

\subsection{Dynamic Survival Analysis}

Following the formulation proposed by \cite{TemporallyConsistentSA} we consider finite sequences of the form $(x_0, \dots, x_{t-1})$ of length $t$ to be drawn from a Markov chain with state space $\cX$ and transition probabilities $p(x'|x)$. 
This assumption allows us to reason about the causality of the data and to leverage temporal consistency\footnote{The Markovian assumption need not be restrictive: depending on the application, states can be formulated to represent the entire past up to that moment.}.
These sequences eventually reach a \emph{terminal state} denoted $\varnothing$, which is absorbing.
The goal of Survival Analysis in this context is to model the time lapse $T$ between the initial state $x_0$ and the event $x_i = \varnothing$ for some $i$.
The random variable $T$, often referred to as \emph{time-to-event}, represents the time at which the event of interest occurs.

Let $F(k|x) = p(T \le k|x_0=x)$ and $f(k|x) = p(T=k|x_0 = x)$ be the distribution function and probability mass function of $T$ respectively, provided the initial state is $x$.
Then we can define two crucial quantities that characterise the survival distribution of $T$.
The \emph{survival function} $S(k|x)$ describes the probability of a sequence starting at state $x$ and \emph{surviving} past a certain time $k$. The \emph{hazard function} $h(k|x)$ represents the risk of a sequence reaching its terminal state at time $k$, provided it departed from the initial state $x$ and that the event of interest has not yet been observed.
They are defined as $S(k|x) = 1 - F(k|x)$
and $h(k|x) = p(T=k|T\ge k, x_0=x)$ respectively.

A very popular choice for modeling survival data is to fit a parametric or semi-parametric form of the hazard function to the observed data,
\begin{equation}
    \label{eq:hazard}
    \text{logit}(h_{\theta}(k|x)) = g_{\theta}(x, k)\,.
\end{equation}
A widely used example of this formulation is the proportional hazards Cox regression model \cite{Cox1972} where $g_{\theta}(x, k) = h_0(k) + \tilde{g}_{\theta}(x)$.
Here, $h_0$ is a \emph{baseline hazard} that depends only on the time and $\tilde{g}_{\theta}$ is a relative risk function that depends only on the covariates.
When handling a discrete time domain one can use the discrete Cox model \cite{allison1982discrete}. In this case the baseline hazard can be treated as a constant or a separate vector of parameters.
Recent literature has expanded this model to incorporate more expressive architectures while keeping the proportional assumption \cite{katzman2018deepsurv}, or alternatively, deviating from it by introducing a non-proportional formulation \cite{Kvamme2019TimetoEventPW}.
Furthermore, one can deviate from the Cox model and parameterise Equation~\eqref{eq:hazard} directly with a neural network \cite{nagpal2021deep,ren2019deep}.

\subsection{Maximum Likelihood Estimation}
The general practice is to collect a dataset with sequences sampled from the Markov chain, each up to a maximum length or \emph{horizon}, denoted $H$.
For some sequences the event of interest is not observed, meaning they do not reach the terminal state. 
These sequences are referred to as \emph{censored}.
Let $\mathbf{1}\{.\}$ be an indicator function that returns $1$ if the condition inside the brackets is true and $0$ otherwise.
Given a sequence $x=(x_0, \dots, x_{t-1})$ of length $t$, 
we define a \emph{censoring indicator} as $c=\mathbf{1}\{x_{t-1} \neq \varnothing\}$,  indicating whether $x$ has reached the terminal state within $t$ steps.
Given a dataset $\cD = \{(x^{(i)}, t^{(i)}, c^{(i)})\}_{i=1}^n$, where $x^{(i)} = (x^{(i)}_0, \dots, x^{(i)}_{t^{(i)}-1})$ is a sequence, $t^{(i)}$ its duration, and
$c^{(i)} = \mathbf{1}\{x_{t-1} \neq \varnothing\}$ is the censoring indicator,
we can frame the survival regression problem as a classification problem \cite{lee2018deephit,craig2021survival}.
We can then express the Maximum Likelihood Estimator (MLE) as follows
\begin{equation}
    \label{eq:MLE}
    \theta^{\star} \in \argmin_{\theta} \sum_{i=1}^n \sum_{k=0}^{H} \sum_{\ell=0}^{k-1} w^{(i)}_{k-\ell} \mathrm{CE}\big(y^{(i)}_{k-\ell},h_{\theta}(k-\ell|x^{(i)}_\ell)\big) \,.
\end{equation}
Here, $\mathrm{CE}$ is the binary cross-entropy loss and $w^{(i)}_k = \mathbf{1}\{t^{(i)} \ge k\}$ is a variable that indicates whether the $i^{\mathrm{th}}$ sequence is still active at time $k$. The event indicator is defined as $y^{(i)}_k = \mathbf{1}\{t^{(i)}-1 = k \wedge c^{(i)} = 0\}$, and indicates whether an uncensored sequence reached the terminal state.
In the dynamic Survival Analysis framework one might fix $\ell=0$. 
In that case the MLE is based on predictions made when departing from the first state, thus fully disregarding intermediate states.
Letting $\ell$ take values larger than zero is an extension called \emph{landmarking} that overcomes this limitation by also regressing the hazard rate of intermediate states \cite{andersen2002multi}. 

\subsection{Temporal Consistency in Survival Analysis}

\cite{TemporallyConsistentSA} introduce Temporally Consistent Survival Regression (TCSR), a method that hinges on the notion of temporal consistency.
The rationale behind this idea is that if we reach the terminal state in $k$ steps when departing from some state $x$, then, on average, that should be equivalent to departing from the next state, and reaching the terminal state after $k-1$ steps.
Temporal consistency makes for a more robust model with respect to noise and helps avoiding overfitting by better capturing the patterns of time-indexed data.
This idea is formally stated in the following identities:
\begin{equation}
    \begin{split}
        f(k|x) &= \E_{x' \sim p(\cdot | x)} \left[ f(k - 1 | x') \right] \\
        S(k|x) &= \E_{x' \sim p(\cdot | x)} \left[ S(k - 1 | x') \right]\,,
    \end{split}
\label{eq:temp_consist}
\end{equation}
where $p(\cdot|x)$ is the transition distribution. 

In a nutshell, the TCSR algorithm works by bootstrapping predictions at the next step from the model's current parameters. These predictions are used as pseudo-targets for updating the model's parameters. A limitation of this approach is that it requires a complete pass through the dataset to compute these soft targets, which can significantly slow down training on large datasets. Additionally, the model must be re-trained from scratch at each step using the updated soft targets, which makes it difficult to process data in batches or perform end-to-end training. The frequent updates of the soft targets also introduce variance in the estimations, leading to unstable training dynamics.

In the original paper, the model's updates are performed using Newton's optimization method at each iteration. The authors prove that the TCSR algorithm is equivalent to a fixed-point iteration problem with the empirical Markov chain and transition probabilities. They also provide a generalization of this algorithm, TCSR($\lambda$), where they introduce the hyperparameter $\lambda$ to control the trade-off between bootstrapping near-term predictions and learning ground truth events from the distant future.

In the following Section we build upon these ideas and provide a mechanism to bring temporal consistency to survival problems involving large datasets with sequences of potentially large horizons.


\section{Proposed Method}
\label{sec:model}

In this Section, we introduce Delayed-network Temporally Consistent Survival Regression (D-TCSR), a temporally-consistent survival regression method that separates the bootstrapping of the training signal from the online parameter updates. D-TCSR leverages a target network that acts as a delayed copy of the main model, introducing a controllable separation between the two update time scales while retaining minibatch, end-to-end optimization.

The idea of a target network mechanism has been widely studied in the Deep RL community, in particular for stabilizing bootstrapped learning in deep TD Learning.
Further details on how our method relates to TD Learning and TCSR are discussed in Appendix~\ref{appendix:connection_dqn}.

\subsection{Scalable Survival Regression with Temporal Consistency}

We consider practical settings where we have access to a dataset of sequences of varying length as described in Section~\ref{sec:preliminaries}.
Our architecture has two main components, a \emph{main network} and a \emph{target network}.
The main network is used to parameterize the hazard function in~\eqref{eq:hazard}, and it has a vector of parameters denoted as $\theta$.
This is the network that will be employed at inference time to calculate the hazard rates using the function $h_{\theta}$.
We also introduce a target network parameterized by $\phi$
that serves as a delayed copy of the main network.
The purpose of the target network is to provide a stable signal for the main network to be updated.
Throughout the training process, given a sequence $x$ with $t$ states\footnote{In this context, one can also say that the sequence $x$ has a duration of $t$ time steps.} and censoring indicator $c$, we use the target network to bootstrap target hazards $h_{\phi}(k|x_{\ell})$ and weights $S_{\phi}(k|x_{\ell})$ for $k = 0, \dots, t - 1$ and $\ell = 0, \dots, t - 1$, based on the current model estimate. These values can be considered the "soft" targets and weights, corresponding to the "hard" value indicators $y_k$ and $w_k$ in Equation~\eqref{eq:MLE}. The supervisory signal is then obtained by interpolating the true "hard" value indicators $y_k$ and $w_k$ from the dataset with the "soft" targets $h_{\phi}(k|x_{\ell})$ and weights $S_{\phi}(k|x_{\ell})$ bootstrapped from the model's target network.
We call these the \emph{pseudo-targets} and \emph{pseudo-weights} respectively.
Only the main network is updated using gradient descent, while the target network is updated via an exponential moving average to remain close to the main network.


\subsubsection{Pseudo-target and pseudo-weight coefficients.}
We supervise the model with pseudo-targets $\ty_{k,\ell}$ and pseudo-weights $\tw_{k,\ell}$ obtained through recursive formulas.
The pseudo-targets are computed as follows:
\begin{equation}
    \label{eq:target}
    \ty_{k,\ell} = \lambda  \ty_{k-1,\ell+1} + (1 - \lambda)  h_{\phi}(k-1|\ell+1)\,,
\end{equation}
where $k, \ell = 0, \dots, t-1$, $k \le \ell$. 
The initialization of the pseudo-targets is done as follows: at the first time step, $k = 0$, we initialize $\ty_{0,\ell} = 0$ because all sequences are considered active, that is, the event of interest is not observed at $k=0$.  
Additionally, when departing from the terminal state, $\ell = t-1$, the pseudo-target is fixed as the true label from the dataset, i.e., $\ty_{k, t-1} = y_{t-1}$.
Using analogous reasoning we derive the recursive formula for the pseudo-weights as follows:
\begin{equation}
    \label{eq:weight}
    \tw_{k,\ell} = \lambda  \tw_{k-1,\ell+1} + (1 - \lambda)  S_{\phi}(k-1|\ell+1)\,,
\end{equation}
where $\tw_{0,\ell} = 1$, $\tw_{k,t-1} = S_{\phi}(k-1|t-1)$ if $c = 1$ and $\tw_{k,t-1} = w_{t-1}$ if $c= 0$.

Intuitively, the equations above compute the pseudo-targets by combining the soft targets and weights bootstrapped from the target network with the "hard" true label indicators and true weight indicators, respectively from the dataset, with the hyperparameter $\lambda$ controlling the trade-off between the two.
When $\lambda = 0$ we only use the soft targets and weights to fit the model.
This implies that we are only taking into account the next state.
In contrast, setting $\lambda=1$ is equivalent to the SA baseline with landmarking \cite{andersen2002multi}.

\subsubsection{Training.}
To fit the model's main parameterizing network we minimize the weighted Cross Entropy loss between the model's estimates and the pseudo-targets. 
The loss function is
\begin{equation}
    \label{eq:loss}
    \cL(\theta) = \sum_{i=1}^n \sum_{\ell=0}^{t^{(i)}-1} \sum_{k=0}^{\ell-1} \tw^{(i)}_{k,\ell} \mathrm{CE}\big(\ty^{(i)}_{k,\ell},h_{\theta}(k-\ell|x^{(i)}_\ell)\big) \,.
\end{equation}

Note that this objective is a weighted MLE with pseudo-targets and weights obtained through \eqref{eq:target} and \eqref{eq:weight}.

We update the main network parameterized by $\theta$ with Stochastic Gradient Descent (SGD).
The target network parameterized by $\phi$ is updated 
with an exponential moving average formula as $\phi = \tau \theta + (1-\tau) \phi$, where $\tau \in (0, 1)$ acts as a target learning rate and is typically close to $0$. 
This ensures the target network will remain close to the main network during training.
Intuitively, a larger value of $\tau$ implies more frequent updates of the target network and hence introduces more variance in the estimations.

The learning procedure is described in Algorithm~\ref{alg:training}, and a direct comparison of the different temporally consistent methods is provided in Appendix~\ref{appendix:comparison}.

\begin{algorithm}[tb]
\caption{D-TCSR algorithm}
\label{alg:training}
\begin{algorithmic}[1]
    \Statex \textbf{Input:}
        $\mathcal{D}$, $\theta$, $\tau$ and $\lambda\in[0,1]$
    \State Initialize $\phi \gets \theta$
    \Repeat
        \State Sample sequence $x$
        \State Compute $h_{\phi}(k\mid x_{\ell})$
            for $k,\ell=0,\dots,t$
        \State Obtain $\ty_{k,\ell}$ and $\tw_{k,\ell}$
            using \eqref{eq:target} and \eqref{eq:weight},
            respectively
        \State Update $\theta$ with SGD by minimizing
            \eqref{eq:loss}
        \State Update $\phi$ with a moving average
    \Until{the maximum number of iterations is reached}
\end{algorithmic}
\end{algorithm}


\section{Experiments}
\label{sec:experiments}

The primary objective of our experimental study is to empirically assess the benefits of introducing a delayed target network in terms of both predictive performance and training stability. 
We first replicate the experimental setup of \cite{TemporallyConsistentSA} in the small-data regime where fitted TCSR can be applied. This allows us to compare fitted TCSR with Inc-TCSR and D-TCSR, using fitted TCSR as a reference for the behavior of a procedure in which target construction and model fitting are more clearly separated.
We then evaluate the same mechanism in larger longitudinal settings, where we instantiate Inc-TCSR and D-TCSR with two established survival models, Logistic-Hazard (LH) \citep{kvamme2021continuous} and DynamicDeepHit (DDH) \citep{lee2019dynamic}. Finally, we study the effect of the target-network update rate \(\tau\) in isolation.

\subsection{Comparing TCSR algorithms in small datasets}
\label{subsec:small_datasets}

In this Section we replicate the experimental protocol introduced by \cite{TemporallyConsistentSA}, and compare fitted TCSR, Inc-TCSR and the delayed target network counterpart, D-TCSR.
We employ a linear model of the form
    $\mathrm{logit}[h_{\theta}(k|x)] = \beta^{\transpose} x + \alpha_k$
with learnable parameters $\theta=(\alpha, \beta)$, for all models.
We conduct experiments on two real-world datasets and one synthetic dataset.

The two real-world datasets come from the medical domain and follow the survival status of patients participating in a specific clinical trial. 
Both combine static features and time-varying features.
In this context, the time-varying or "dynamic" features are biomarkers that are measured periodically within a time window of many years.
For our experiments we only considered events occurring at discrete time points, thus disregarding the potentially different time intervals between each pair of measurements.
The first dataset, PBC2, follows 312 patients suffering from a liver disease called Primary Biliary Cirrhosis \cite{murtaugh1994primary}.
The second dataset, AIDS, contains information of 467 patients diagnosed with HIV/AIDS \cite{abrams1994comparative}.
We finally consider a synthetic dataset where we generated sequences using a $20$-dimensional Gaussian random walk (SmallRW) with the horizon set to $11$.
At each step $k$ we emulated a churn event as follows: the sequence finishes with probability $p = \sigma(a^{\transpose} x_k + b)$ where $a\in\realset^{20}$ and $b\in\realset$ are user-input hyperparameters.

\begin{figure}[h]
    \centering
    \includegraphics[width=\textwidth]{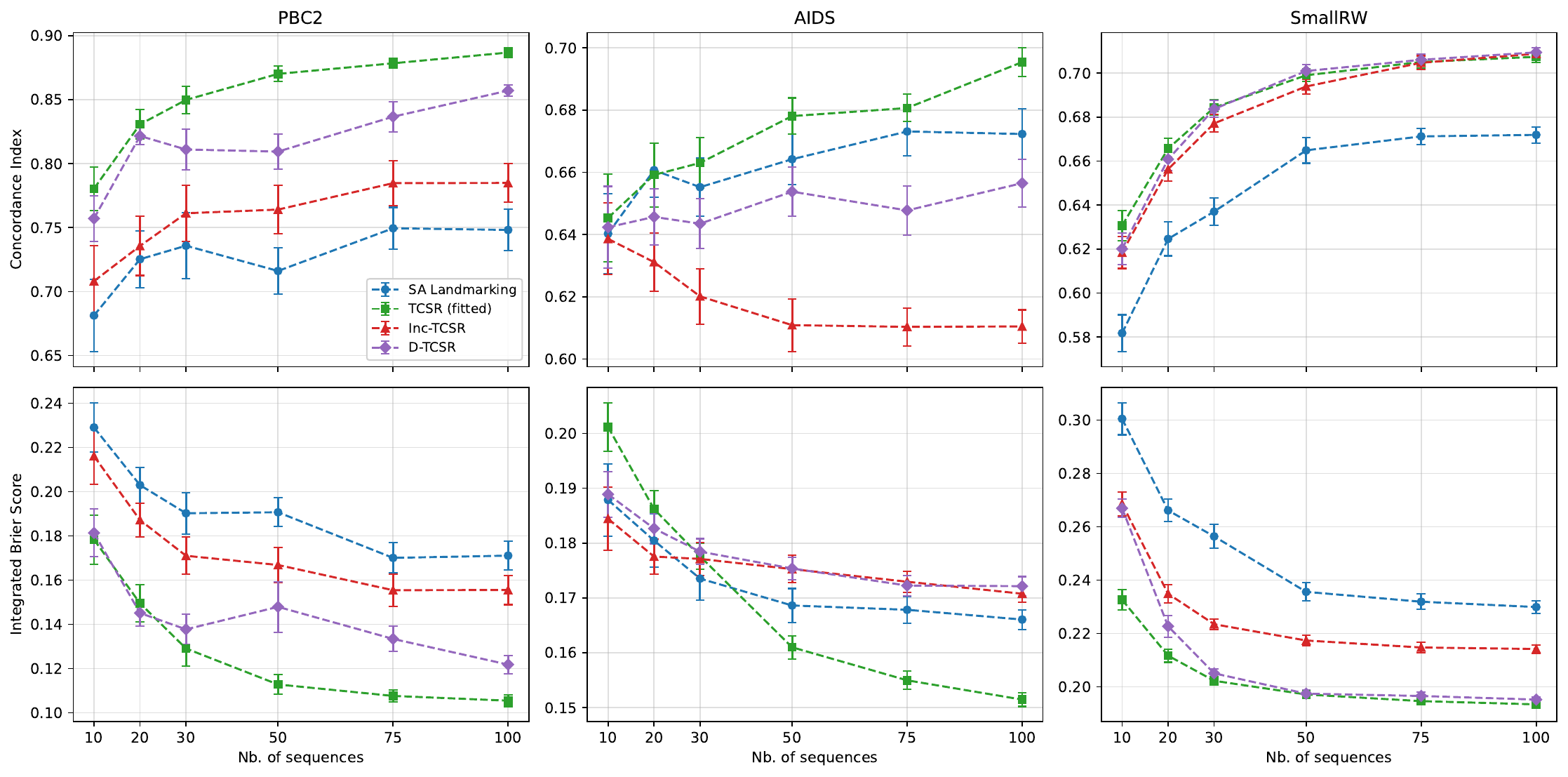}

    \caption{Performance of event prediction in small datasets. We present results in terms of the CI (higher is better) and IBS (lower is better). The error bars represent the mean and standard deviation over thirty random seeds.}

    \label{fig:res_small}
\end{figure}

We compare fitted TCSR, Inc-TCSR and D-TCSR with a baseline consisting of regular Survival Analysis with landmarking (SA Landmarking).
For all TCSR algorithms we fix $\lambda = 0$.
Inc-TCSR is the $\tau = 1$ instance and therefore has no hyperparameter to tune. 
For D-TCSR we tune only $\tau$, over $\{0.001, 0.003, 0.01, 0.05, 0.1, 0.25, 0.5, 0.75, 0.9, 0.95, 0.99\}$, whereas for
fitted TCSR we tune the $\ell_2$ penalty over $\{0, 0.01, 0.1, 1\}$ (100 outer iterations). 
We split each dataset using $5$-fold cross validation and fit the models using training sets with size in $\{10, 20, 30, 50, 75, 100\}$.
Each training set size uses a test set fixed across all runs (20\% of the sequences). 
The remaining sequences are reshuffled per seed into a training pool and a validation set
with a \%75/\%25 proportion, and the training pool is truncated to its first $n$ sequences.
Hyperparameters are selected on validation, separately for each metric: the
configuration with the highest mean validation C-index over seeds 0--9 is
reported for the C-index, and the one with the lowest mean validation IBS for
the IBS. 
Selection is done once per (method, dataset, $n$), and the selected configuration is then trained on 20 additional seeds.
For PBC2 and SmallRW we fix the weight decay regularization hyperparameter to $0.0$, whereas for AIDS we fixed it to $0.4$.
We set the SGD learning rate to $0.1$, and we fit the model during 100 epochs for all datasets with Adam optimizer \citep{kingma2014adam}.
We report results using two widely used SA metrics: average Concordance Index (CI) and Integrated Brier Score (IBS).
Averaged values across 30 random splits ($\pm$ one standard error) are summarized in Figure~\ref{fig:res_small}.

Overall, temporally consistent methods outperform the landmarking baseline across training-set sizes. 
Among the bootstrapped variants, D-TCSR consistently improves over Inc-TCSR on PBC for both metrics, and shows additional gains on AIDS in terms of CI and on SmallRW in terms of IBS.
On SmallRW, the three temporally consistent methods perform similarly in terms of CI. 
Fitted TCSR remains the strongest method in general. 
This is consistent with its optimization procedure: within each outer iteration, the pseudo-targets are held fixed while the survival model is fitted, fully separating target construction from parameter updates. 
Inc-TCSR removes this separation by updating the model against targets generated from the same rapidly changing parameters, whereas D-TCSR only partially restores it through a slowly updated target network. 
The results therefore suggest that delaying the targets recovers part of the benefit of fitted target separation while retaining stochastic optimization.

\subsection{Extending survival algorithms with temporal consistency}
\label{subsec:large_datasets}

\subsubsection{Baselines and setup} 
Following the landmark-based evaluation protocol of \cite{lee2019dynamic}, we evaluate predictions at multiple landmark times \(t\) and prediction horizons \(\Delta\) using the time-dependent concordance index \(C(t,\Delta)\) and the inverse-probability-of-censoring-weighted (IPCW) Brier score \(\mathrm{BS}(t,\Delta)\) (further details on these metrics are provided in Appendix~\ref{appendix:timedep_ev_metrics}). 
Landmarks and prediction horizons form a grid that is picked once for each dataset and shared for all models and seeds. 
The landmarks are chosen as the 25th, 50th and 75th percentiles of the observed times (event or censoring) of all subjects.  
At each landmark, the prediction horizons are the 25th, 50th and 75th percentiles of the remaining time $T^{(i)} - t$ among the subjects still at risk. Percentiles are rounded down to whole steps, and duplicate horizons are merged.
In this experimental protocol, landmarks probe early, middle and late points of the follow-up, and its
horizons probe short, medium and long windows relative to each subject's remaining lifetime. 

We evaluate two well established SA models.
The first baseline is the Logistic-Hazard (LH) model \cite{kvamme2021continuous} for time-varying covariates described in~\eqref{eq:hazard}, where we parameterize the hazard function with a neural network. The second model we include in our experiments is DynamicDeepHit (DDH) \cite{lee2019dynamic}, a deep learning-based survival analysis model that jointly estimates the probability of survival and the occurrence of competing events over time in longitudinal data.
We compare each baseline model, $A$ baseline, $A\in\{\text{LH, DDH}\}$, against the proposed temporally-consistent counterparts, namely $A$ Inc-TCSR and $A$ D-TCSR. 
For the temporally consistent variants, the "hard" targets from the training set are replaced by "soft" targets as described in Section~\ref{sec:model} by the means of a target network.

For both model families, hyperparameters are selected in two stages. 
We first select a common architecture for each family, shared across the baseline, Inc-TCSR, and D-TCSR arms, and then tune the temporal-consistency hyperparameters while keeping the architecture fixed. 
Inc-TCSR corresponds to \(\tau=1\).
For both temporally consistent variants we tune \(\lambda\), while D-TCSR additionally tunes \(\tau<1\). 
We select hyperparameters separately for each dataset and evaluation metric using the mean validation score over seeds 0--9, averaged across landmark--horizon cells. 
We select a configuration per (method, dataset, metric) which is then evaluated over 30 seeds, and we report the mean test performance with one standard error and paired \(t\)-tests across seeds. 
Full architecture, optimization, and hyperparameter-search details are provided in Appendix~\ref{appendix:arch_details}.

\subsubsection{Datasets} Our first dataset is LastFM, a churn prediction dataset \cite{Bertin-Mahieux2011}.
Churn prediction is the task of detecting which custumers are likely to stop using a service. 
Provided a set of user features or logs associated with "churning times", it is fairly straightforward to formulate churn prediction as a survival problem.
We use a subset of the full LastFM dataset containing the listening history of almost 1000 users between 2004 and 2009 in a popular music streaming platform.
Specific details about the times when users discontinued the service are also included.
Users who were inactive in the platform for a week prior to the ending date of the study are considered to have churned. 
In this work we focus exclusively on the time-varying features, which consists of
 time-stamped logs that specify which artist and song the user is listening to.
 We aggregate events into weekly windows and extract, for each user-week, a set of behavioral features capturing engagement and diversity: total listen count, number of unique artists, artist entropy (Shannon entropy over the artist distribution within the window), repeat listen ratio (fraction of plays that are re-listens), and mean song novelty (defined as the inverse of cumulative prior listens for each song at the time of play, averaged over the window). To capture longitudinal trends, we additionally compute week-over-week deltas for each of these features.

We also include a dataset collected by \cite{saxena2008damage} consisting of simulated measurements from aircraft engines that run until failure.
It provides time-series measurements of several engine parameters overtime, allowing for the analysis of the engine degradation process.
The event of interest is engine failure, and censored sequences correspond to engines that did not fail during the study period.
Approximately 50\% of the sequences are censored, and the dataset has an average horizon of 168 time steps, with some extending beyond 300 time steps, making it a large horizon dataset.

An additional benchmark is provided by the public SCANIA Component X dataset \citep{Kharazian2025} (SCANIA), released for
  the Symposium on Intelligent Data Analysis (IDA) 2024 industrial challenge. 
  This dataset contains real operational
  data from a fleet of Scania trucks: for each vehicle, a sequence of irregularly
  sampled readouts of 105 anonymized cumulative counters, together with the
  length of its study period and whether an anonymized engine component (``X'')
  was repaired during it. 
  These counteres are log-transformed as $\log(1+x)$ to compress its range, and then standarized to zero mean and unit variance.
  Vehicles without a repair are right-censored. 
  The training release covers 23{,}550 vehicles (1.12M readouts, 9.6\% repaired). 
  We use a fixed random subsample of 5{,}000 vehicles (475 repairs, 90.5\%
  censored), index time by readout rather than calendar time, truncate sequences
  at 100 readouts and treating longer ones as censored at the horizon.

In the current literature the availability of mid and large scale survival datasets is very limited.
While in general we can find large datasets in the context of static SA, the scarcity of datasets is particularly pronounced in the context of time-varying settings.
For this reason we complete this experimental section with a large-scale synthetic dataset (LargeRW).
We used a Gaussian random walk model where, 
at each step $k$, we emulated a churn event as follows: the sequence finishes with probability $p = \sigma(a^{\transpose} x_k + b)$ where $a\in\realset^{20}$ and $b\in\realset$ are user-input hyperparameters.
The generated data set has 50 features and a horizon $H=100$.
We generated 10{,}000 samples with about 22\% of censored sequences.
Table~\ref{datasets} presents summary statistics for all datasets.

\begin{table}[t]
\caption{Statistics of all datasets.}
\label{datasets}
\begin{center}
\begin{tabular}{lcccr}

\bf DATASET  & \bf SIZE & \bf \#STATIC & \bf \#TIME-VAR. & $H$\\
\hline \\
PBC2      & 312 & 3 & 12 & 16\\
AIDS      & 467 & 4 & 1  & 5 \\
SmallRW   & \--- & 0& 20 & 11\\
LargeRW   & 10k & 0& 50 & 100\\
LastFM    & 992 & 0 & 10 & 226\\
NASA      & 200 & 0 & 16 & 361 \\
SCANIA     & 5000 & 0 & 105 & 100 \\

\end{tabular}
\end{center}
\end{table}

\subsubsection{Results} 

We report results in terms of "number of wins" per \emph{cell}.
In this context, a \emph{cell} is one $(t, \Delta)$ pair, where $t$ is a landmark and $\Delta$ is a predictive horizon. 
  In each cell we score the test subjects at risk at $t$ with $C(t,\Delta)$ and $\mathrm{BS}(t,\Delta)$. 
  Each method uses the single configuration selected
  for its dataset and metric, averaged over the 30
  seeds.
  Cells with fewer than 10 such subjects
  are skipped, and seeds for which $C(t,\Delta)$  has no comparable
  pairs are dropped for that cell only. 
We say that a method "wins" a cell
  if it has the best mean metric there, and the win is significant when it
  beats both other methods at $p < 0.05$ according to a paired $t$-test across seeds. 
  With three landmarks and up to three
  horizons each, every dataset contributes nine cells, except LargeRW, where
  duplicate horizons at the last landmark leave seven. 

Results for the LH and DDH families are summarized in Tables~\ref{tab:lh_wins} and~\ref{tab:ddh_wins}, respectively. 
For LH, D-TCSR is the best-performing method in 61 of 68 landmark--horizon cells, with 36 statistically significant wins. 
More importantly, in the direct comparison with Inc-TCSR, D-TCSR performs better in 62 of 68 cells and significantly so in 54. 
A similar pairwise pattern emerges for DDH: although the untreated DDH baseline is more competitive and D-TCSR is the overall winner in 42 of 68 cells, D-TCSR still outperforms Inc-TCSR in 63 cells, with 54 significant improvements. 
These results support the hypothesis that separating the update timescales of the online model and the bootstrap target can improve temporally consistent training, even in settings where temporal consistency itself does not necessarily improve over the underlying survival model. 
Fine-grained results for \(C(t,\Delta)\) and \(\mathrm{BS}(t,\Delta)\) across all landmark--horizon cells are reported in Appendix~\ref{app:fine_grained_results}.

\begin{table}[t]
\caption{
Results across landmark--horizon configurations for LH.
For the three-way comparison, entries report the number of configurations
won by each method, with the number of significant wins in parentheses.
A significant win means that the winning method outperforms both other
methods in paired $t$-tests over 30 seeds ($p<0.05$).
The final column reports the direct comparison between D-TCSR and
Inc-TCSR as \emph{better (significant)}.
Higher $C(t)$ and lower Brier score are better.
}
\label{tab:lh_wins}
\begin{center}
\begin{tabular}{llrrrrrr}
\multicolumn{1}{c}{\bf METRIC} &
\multicolumn{1}{c}{\bf DATASET} &
\multicolumn{1}{c}{\bf CELLS} &
\multicolumn{1}{c}{\bf LH} &
\multicolumn{1}{c}{\bf Inc-TCSR} &
\multicolumn{1}{c}{\bf D-TCSR} &
\multicolumn{1}{c}{\bf D $>$ Inc} 
\\ \hline \\

$C(t)$
& NASA     & 9 & 0 (0) & 0 (0) & \textbf{9 (5)} & \textbf{9 (9)}  \\
& Scania   & 9 & 0 (0) & 0 (0) & \textbf{9 (4)} & \textbf{9 (9)} \\
& LastFM   & 9 & 0 (0) & 0 (0) & \textbf{9 (7)} & \textbf{9 (7)}  \\
& Large-RW & 7 & 0 (0) & 0 (0) & \textbf{7 (7)} & \textbf{7 (7)}  \\
& \textbf{Total}
& \textbf{34}
& \textbf{0 (0)}
& \textbf{0 (0)}
& \textbf{34 (23)}
& \textbf{34 (32)}

\\\hline 

Brier
& NASA     & 9 & 1 (0) & 0 (0) & \textbf{8 (5)} & \textbf{9 (9)}  \\
& Scania   & 9 & 0 (0) & 0 (0) & \textbf{9 (3)} & \textbf{9 (7)}  \\
& LastFM   & 9 & 0 (0) & 1 (0) & \textbf{8 (5)} & \textbf{8 (6)}  \\
& Large-RW & 7 & \textbf{4 (3)} & 1 (0) & 2 (0) & 2 (0)  \\
& \textbf{Total}
& \textbf{34}
& \textbf{5 (3)}
& \textbf{2 (0)}
& \textbf{27 (13)}
& \textbf{28 (22)}

\\\hline 

\multicolumn{2}{l}{\bf Overall}
& \textbf{68}
& \textbf{5 (3)}
& \textbf{2 (0)}
& \textbf{61 (36)}
& \textbf{62 (54)}
\end{tabular}
\end{center}
\end{table}

\begin{table}[t]
\caption{
Results across landmark--horizon configurations for DDH.
For the three-way comparison, entries report the number of configurations
won by each method, with the number of significant wins in parentheses.
A significant win means that the winning method outperforms both other
methods in paired $t$-tests over 30 seeds ($p<0.05$).
The final column reports the direct comparison between D-TCSR and
Inc-TCSR as \emph{better (significant)}.
Higher $C(t)$ and lower Brier score are better.
}
\label{tab:ddh_wins}
\begin{center}
\begin{tabular}{llrrrrrr}
\multicolumn{1}{c}{\bf METRIC} &
\multicolumn{1}{c}{\bf DATASET} &
\multicolumn{1}{c}{\bf CELLS} &
\multicolumn{1}{c}{\bf DDH} &
\multicolumn{1}{c}{\bf Inc-TCSR} &
\multicolumn{1}{c}{\bf D-TCSR} &
\multicolumn{1}{c}{\bf D $>$ Inc} 
\\ \hline \\

$C(t)$
& NASA     & 9 & 0 (0) & 0 (0) & \textbf{9 (9)} & \textbf{9 (9)} \\
& Scania   & 9 & \textbf{8 (2)} & 0 (0) & 1 (0) & \textbf{8 (6)}  \\
& LastFM   & 9 & 3 (1) & 0 (0) & \textbf{6 (4)} & \textbf{8 (8)}  \\
& Large-RW & 7 & 2 (2) & 0 (0) & \textbf{5 (2)} & \textbf{7 (4)} \\
& \textbf{Total}
& \textbf{34}
& \textbf{13 (5)}
& \textbf{0 (0)}
& \textbf{21 (15)}
& \textbf{32 (27)}
\\ \hline

Brier
& NASA     & 9 & 0 (0) & 0 (0) & \textbf{9 (9)} & \textbf{9 (9)}  \\
& Scania   & 9 & \textbf{7 (3)} & 0 (0) & 2 (0) & \textbf{9 (5)}  \\
& LastFM   & 9 & 3 (1) & 0 (0) & \textbf{6 (1)} & \textbf{9 (9)}  \\
& Large-RW & 7 & 1 (1) & 2 (0) & \textbf{4 (4)} & \textbf{4 (4)}  \\
& \textbf{Total}
& \textbf{34}
& \textbf{11 (5)}
& \textbf{2 (0)}
& \textbf{21 (14)}
& \textbf{31 (27)}
\\ \hline

\multicolumn{2}{l}{\bf Overall}
& \textbf{68}
& \textbf{24 (10)}
& \textbf{2 (0)}
& \textbf{42 (29)}
& \textbf{63 (54)}
\end{tabular}
\end{center}
\end{table}

\subsection{Stabilizing effect of the target network and the choice of $\tau$}
\label{subsec:stabilizing_tau}

The stabilizing capabilities of introducing a target network have been widely studied and demonstrated in the context of Deep Reinforcement Learning \citep{mnih2015humanlevel,lillicrap2016continuous,liang2018categorical}.
To explore the impact of our method in the variability of the estimates in the context of Survival Analysis, we perform an ablation study on the target learning rate $\tau$.
Furthermore, we also investigate the effect of different target learning rates on the performance of the models in terms of CI and IBS, demonstrating the benefits of less frequent updates of the pseudo-targets, as introduced by our framework.

 \begin{table}[t]
  \caption{Ablation study over the target learning rate $\tau$ for random-walk
  datasets with horizons $H \in \{30, 50, 100\}$. We report the mean and
  standard deviation over 30 runs of CI (higher is better) and IBS (lower is
  better). For each horizon and metric, the best mean, and every value not
  significantly different from it according to a paired $t$-test with
  significance level $\alpha = 0.05$, is highlighted in bold.}
  \label{table:ablation_tau}
  \begin{center}
  \small
  \begin{tabular}{llllllll}
  \multicolumn{1}{c}{\bf $H$} & \multicolumn{1}{c}{\bf METRIC}
  & \multicolumn{1}{c}{\bf $\tau = .01$} & \multicolumn{1}{c}{\bf $\tau = .05$}
  & \multicolumn{1}{c}{\bf $\tau = .1$}  & \multicolumn{1}{c}{\bf $\tau = .25$}
  & \multicolumn{1}{c}{\bf $\tau = .5$}  & \multicolumn{1}{c}{\bf $\tau = 1.0$}
  \\ \hline \\
  \multirow{2}{*}{30}  & CI$\uparrow$    & .785 $\pm$ .007 & .786 $\pm$ .007 & .803 $\pm$ .005 & \textbf{.805 $\pm$ .003} & .802 $\pm$ .004 & .795 $\pm$ .013 \\
                       & IBS$\downarrow$ & .216 $\pm$ .009 & .152 $\pm$ .003 & .140 $\pm$ .003 & \textbf{.139 $\pm$ .002} & .142 $\pm$ .003 & .149 $\pm$ .008 \\
  \hline \\
  \multirow{2}{*}{50}  & CI$\uparrow$    & .760 $\pm$ .012 & .771 $\pm$ .008 & \textbf{.788 $\pm$ .004} & \textbf{.789 $\pm$ .003} & .782 $\pm$ .005 & .766 $\pm$ .011 \\
                       & IBS$\downarrow$ & .261 $\pm$ .005 & .162 $\pm$ .004 & \textbf{.150 $\pm$ .002} & \textbf{.149 $\pm$ .002} & .156 $\pm$ .004 & .173 $\pm$ .015 \\
  \hline \\
  \multirow{2}{*}{100} & CI$\uparrow$    & .682 $\pm$ .018 & .712 $\pm$ .006 & \textbf{.726 $\pm$ .003} & .711 $\pm$ .005 & .701 $\pm$ .017 & .659 $\pm$ .044 \\
                       & IBS$\downarrow$ & .285 $\pm$ .007 & .192 $\pm$ .005 & \textbf{.172 $\pm$ .002} & .175 $\pm$ .002 & .185 $\pm$ .008 & .199 $\pm$ .014 \\
  \end{tabular}
  \end{center}
  \end{table}

To that end, we use three Gaussian random walk datasets with different horizons.
Specifically, we generated three $20$-dimensional random walk datasets as described in Section~\ref{subsec:large_datasets} with horizons $H=30$, $H=50$ and $H=100$, adjusting the churn bias so that all datasets have approximately 20\% of censored sequences.
For all datasets, we fix $\lambda = 0.1$, the value selected for D-TCSR on LargeRW, and train models for different values of $\tau$, $\tau\in\{0.01, 0.05, 0.1, 0.25, 0.5, 1.0\}$, where $\tau = 1$ corresponds to Inc-TCSR.
For each horizon and value of $\tau$ we fit LH models on thirty training sets of $1000$ sequences, each randomly generated with a different seed together with a validation set of $500$ sequences. 
All models share the same initialization, so that the variability across runs is due to the training data.
Training was conducted for up to 1000 epochs, with early stopping (patience 5) on the validation loss, using the same architecture and optimizer as described in Section~\ref{subsec:large_datasets} for LargeRW.
Unlike the landmark evaluation used in Section~\ref{subsec:large_datasets}, here models are evaluated from the initial state of each sequence, as in \cite{TemporallyConsistentSA}: on a separate test set of $1000$ sequences, we compute the CI of the predicted median survival time and the IBS of the predicted survival curve.

For each sequence in the test set we collect the estimated hazard probabilities $h_{k\ell}=h_{\theta}(k|x_{\ell})$ for $k=1, \dots, t$ and $\ell=1, \dots, t$ obtained by each model.
We then calculate the mean and standard deviation of the obtained hazard probabilities, $\bar{h}_{k\ell}$ and $\Delta h_{k\ell}$ respectively. 
To obtain the variability of the estimates we re-scale the standard deviations using the standard deviation of the Bernoulli distribution with parameter $\bar{h}_{k\ell}$.
This re-scaling has the purpose of “canceling out” the inherent variability of the hazard distribution. 
That is, we calculate the variability of the estimations as 

\begin{equation*}
    \delta_{k\ell} = \frac{\Delta h_{k\ell}}{\sqrt{\bar{h}_{k\ell}(1 - \bar{h}_{k\ell})}}\,.
\end{equation*}

\begin{figure}[h]
    \centering

    \begin{minipage}{\textwidth}
        \centering
        \includegraphics[width=.8\textwidth,keepaspectratio]{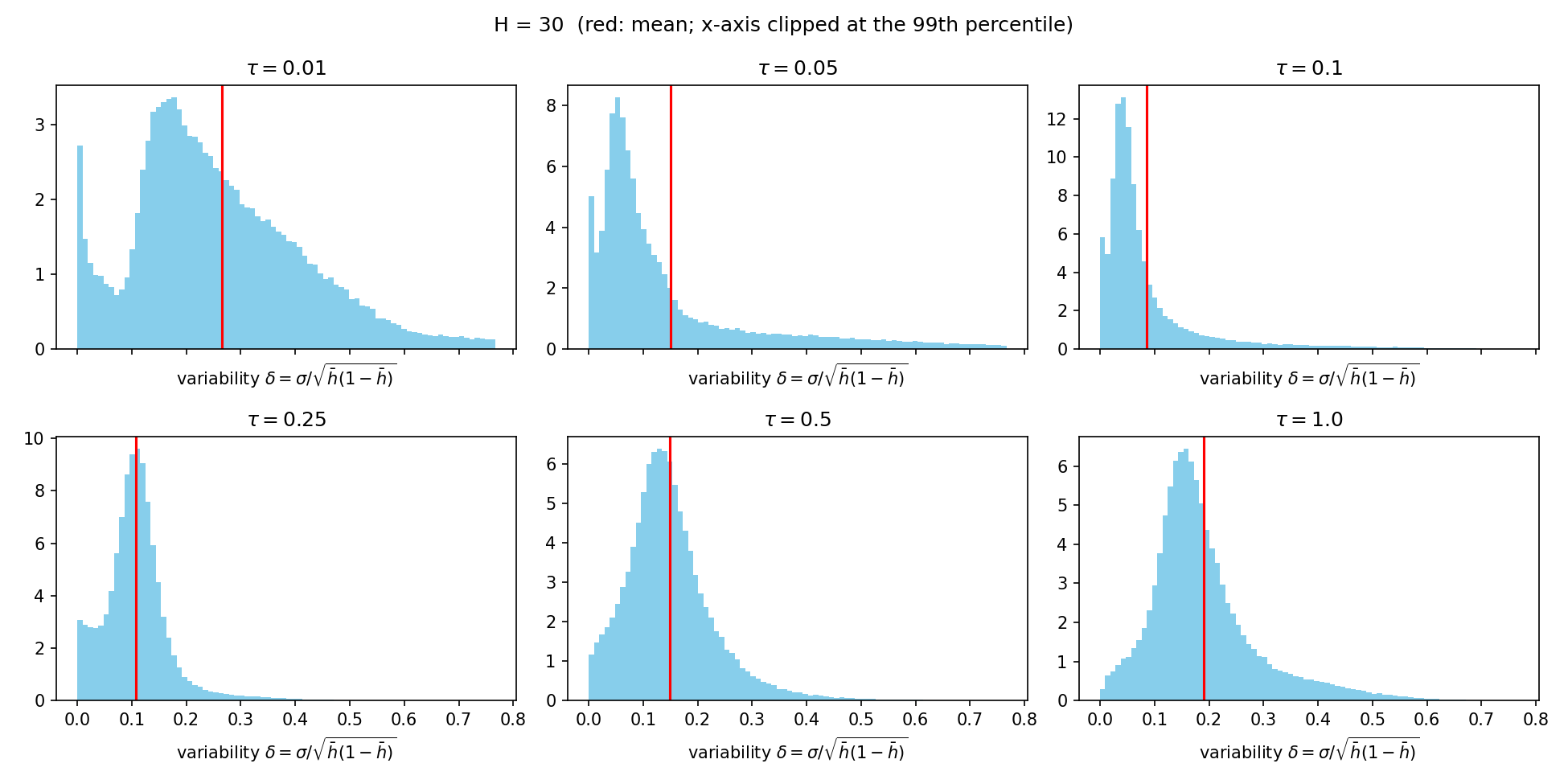}
        \parbox{\textwidth}{\centering (a) $H=30$}
    \end{minipage}

    \vspace{1em} 

    \begin{minipage}{\textwidth}
        \centering
        \includegraphics[width=.8\textwidth,keepaspectratio]{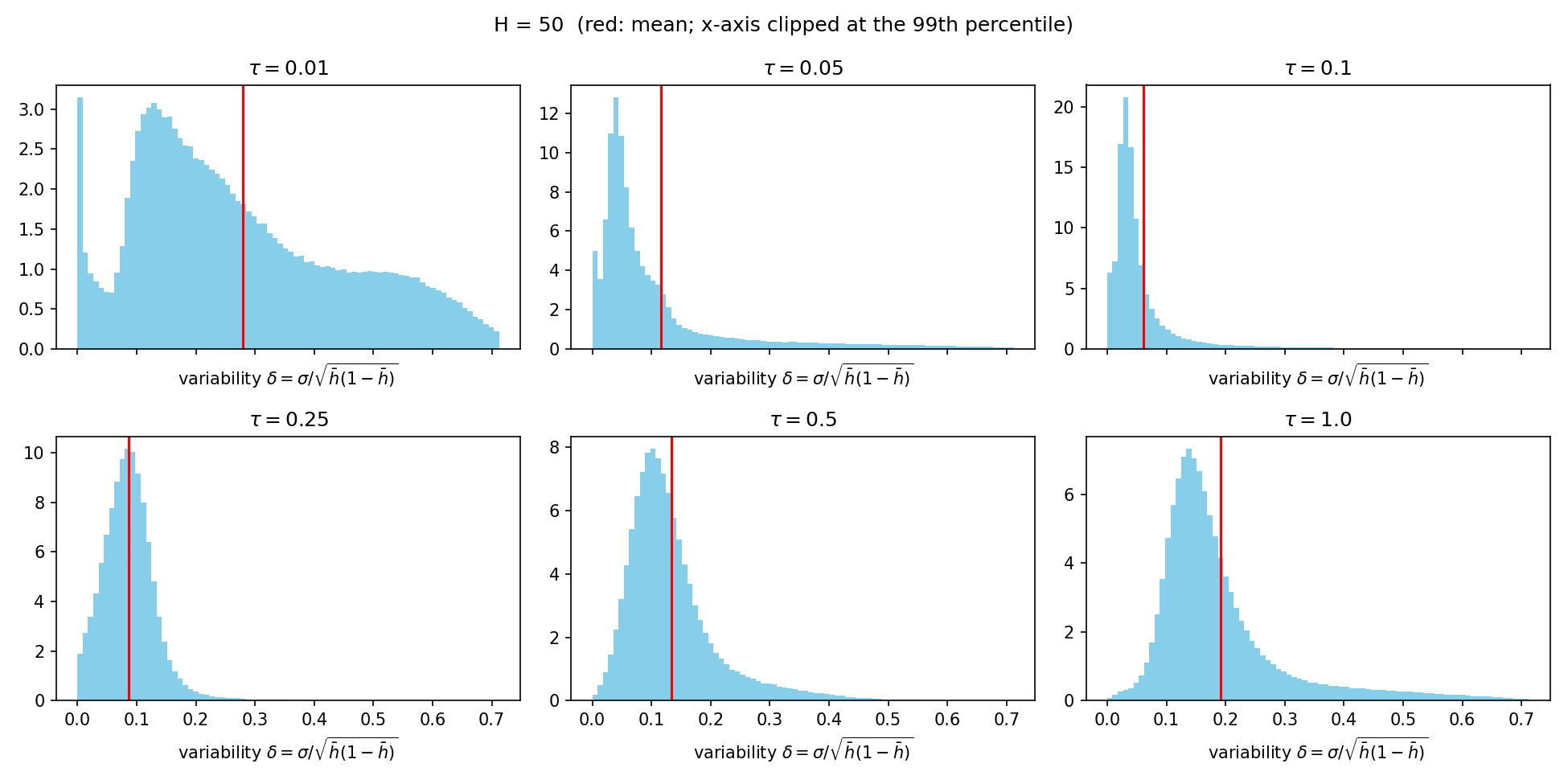}
        \parbox{\textwidth}{\centering (b) $H=50$}
    \end{minipage}

    \vspace{1em} 

    \begin{minipage}{\textwidth}
        \centering
        \includegraphics[width=.8\textwidth,keepaspectratio]{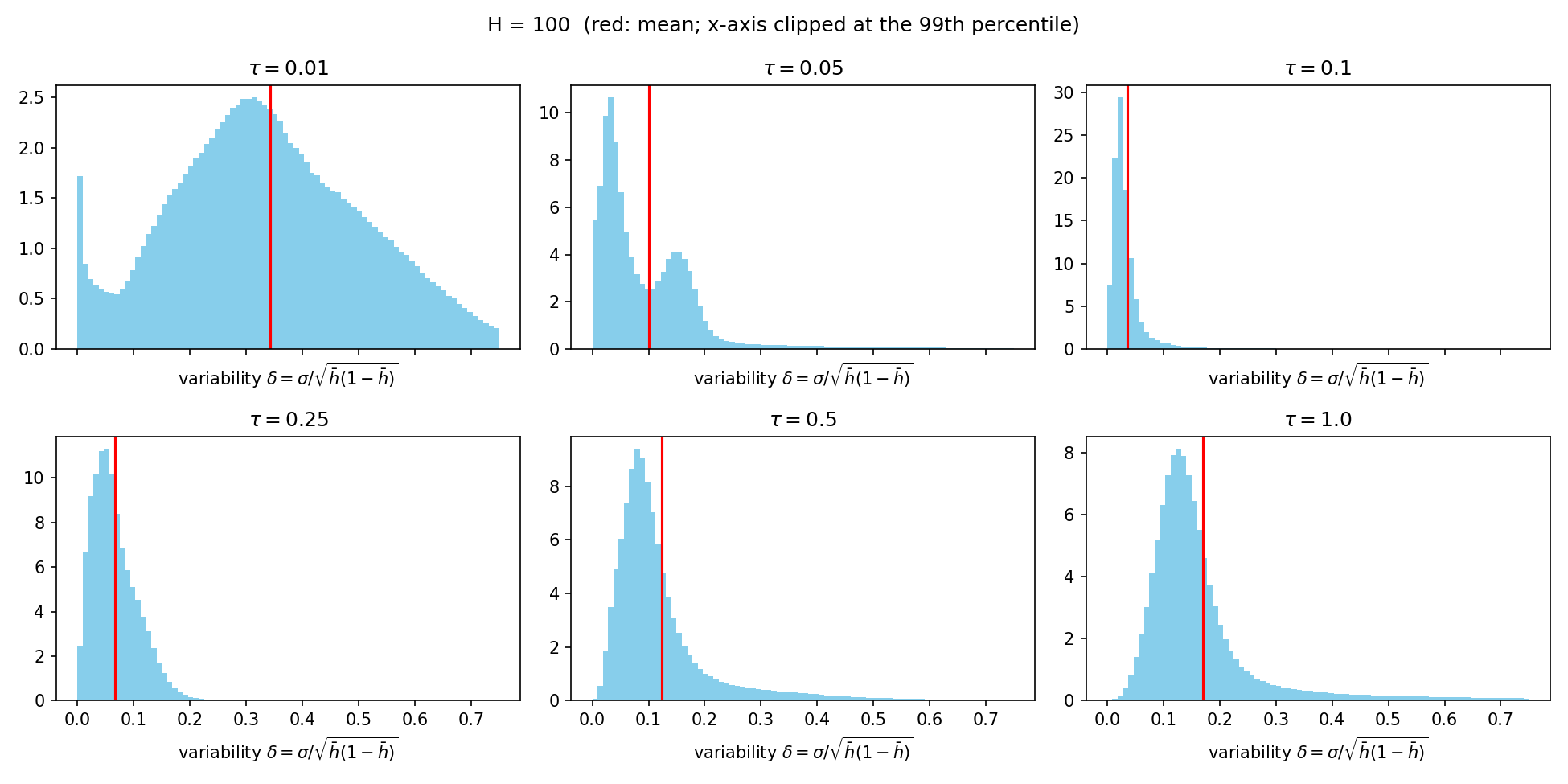}
        \parbox{\textwidth}{\centering (c) $H=100$}
    \end{minipage}

    \caption{Variability of estimates for different values of $\tau$ with $\lambda = 0.1$ for datasets with varying horizons: (a) $H=30$, (b) $H=50$, and (c) $H=100$. The red line indicates the mean value.}
    \label{fig:density_plots}
\end{figure}

In Figure~\ref{fig:density_plots} we present histograms of the variability of the estimates $\delta$ over the test set for the different values of $\tau$ and datasets with horizons $H=30, 50, 100$.
For all horizons, the variability of the estimates decreases as $\tau$ goes from $1$ down to $0.1$, where it reaches its minimum (the mean $\delta$ drops from $0.19$ to $0.09$ for $H=30$, from $0.19$ to $0.06$ for $H=50$, and from $0.17$ to $0.04$ for $H=100$).
This indicates that less frequent updates of the pseudo-targets introduced by the target network reduce the variability of the estimates, thus stabilizing the training procedure.
For $\tau \leq 0.05$ the variability increases again, as the target network lags too far behind the online network and the pseudo-targets become stale.
We present the average performance of the models on the test set for different values of $\tau$ in Table~\ref{table:ablation_tau} for all datasets. Outperforming scores are highlighted based on a paired t-test with significance level $\alpha=0.05$.
For all horizons, the best CI and IBS are achieved with $\tau < 1$, meaning pseudo-targets are updated less frequently, and they are significantly better than those of $\tau = 1$.
The best performing values cluster around $\tau=0.1$ and $\tau=0.25$ for all horizons.
This stands in contrast with the Inc-TCSR method of \cite{TemporallyConsistentSA}, which updates pseudo-targets at every iteration (i.e., $\tau=1$).
The stabilizing effect is also visible in the spread of the scores across training sets: for $H=100$, the standard deviation of the CI drops from $0.044$ for $\tau=1$ to $0.003$ for $\tau=0.1$.
Performance gains are particularly pronounced for larger horizons, where the best $\tau$ values improve over $\tau=1$ by about 10\% in CI and 14\% in IBS for $H=100$, and by about 3\% in CI and 14\% in IBS for $H=50$.
For $H=30$, while smaller $\tau$ values still significantly outperform $\tau=1$, the margin is narrower.
Conversely, very small values of $\tau$ degrade performance: $\tau = 0.01$ yields the worst IBS for all horizons, as the target network moves too slowly to propagate the information in the pseudo-targets within the training budget, so they become stale.
This trend suggests that the choice of $\tau$ reflects a trade-off between stability and staleness, and that the benefits of controlling the frequency of pseudo-target updates become more pronounced as the prediction horizon increases.

\section{Conclusion and Limitations}
\label{sec:conclusion}

We introduced a novel framework for Survival Analysis, designed to efficiently handle time-to-event data with time-varying covariates at scale. Drawing inspiration from the DQN algorithm, our approach introduces a target network that enables temporal consistency in settings requiring mini-batch updates. By separating the model and training-signal update timescales, our method can accommodate complex architectures and end-to-end training while retaining the benefits of temporal consistency. Moreover, we showcased the practical benefits of our approach by extending two representative survival models with temporal consistency.

Across both model families, our experiments show that delaying the bootstrapping of the targets consistently improves over the same-timescale updates of Inc-TCSR. These benefits persist across datasets, architectures, landmarks and prediction horizons. Furthermore, our ablations on $\tau$ show that the amount of target delay is an important component of the method. In particular, intermediate update rates tend to outperform not only Inc-TCSR but also excessively slow target updates, suggesting a trade-off between moving and stale targets.

Our approach also has limitations.
Maintaining a separate copy of the main model increases the memory requirements relative to the underlying survival model. Also, the current formulation assumes a survival model that explicitly parameterizes a discrete hazard function or time-to-event distribution, making it unsuitable for other estimation principles.
Extending temporal consistency to broader classes of survival estimators is a natural direction for future work.

\bibliography{main}
\bibliographystyle{tmlr}

\newpage


\appendix

\section{Connection to TCSR and DQN}
\label{appendix:connection_dqn}

In Reinforcement Learning, Temporal Difference learning with $\lambda$ return (TD($\lambda$)) \cite{sutton1988learning} is a learning algorithm that aims at estimating a scalar-valued function that represents the expected reward based on accumulated experiences over time.
The $\lambda$ coefficient captures a sense of memory over a time window. 
Intuitively, we update our estimates by taking into account a certain range of future events.
When $\lambda$ is closer to $0$ we privilege more recent events, in contrast, when $\lambda$ approaches $1$, we place greater emphasis on events that occurred further in the future.
TCSR and DeepTCSR utilize TD($\lambda$) as a foundation, exploiting the consistency assumption described by Equations~\eqref{eq:temp_consist}. 

On the other hand, Q-Learning \cite{watkins1992q} is a particular instance of TD learning where the goal is to estimate a scalar function that maximizes the expected return of all possible transitions when departing from a specific initial state.
The Deep Q-Network (DQN) algorithm \cite{mnih2013atari} combines Q-learning with deep neural networks in order to handle large-scale state spaces and leverage complex model architectures.
It employs a technique called experience replay, which consists of storing past transitions in memory. 
These transitions are sampled randomly during training, providing a diverse set of experiences for the network to learn from. 
Experience replay is a form of using past temporal differences to update the model.
To stabilize training, DQN introduces a target network that essentially is a delayed copy of the main Q-network.
This target network is used to generate target values for training the model. 
In other words, the model is supervised with soft labels or "slow moving labels".

In the same way that DQN brought deep learning and scalable solutions to Q-learning, our approach extends TCSR by making temporal consistency applicable to scenarios where mini-batch updates and complex architectures are needed, while retaining training stability. 
Unlike Inc-TCSR, which directly updates the training targets at every step, we compute soft targets using the target network.
Intuitively, we can control the update frequency of the target network through the hyperparameter $\tau$.
This not only provides a more consistent and robust signal for parameter updates, but also has a positive impact on predictive performance, as shown in Section~\ref{subsec:large_datasets}. 
This is analogous to how DQN uses experience replay's past transitions to ensure more stable updates.
This separation of the target network from the main model also enables end-to-end training, allowing us to leverage more expressive neural architectures to model complex temporal patterns. 


\section{Further Details of TCSR}
\label{appendix:comparison}

In Table~\ref{tab:comparsion_algs} we summarize the main hyperparameters of the SA baselines with and without landmarking, and for the temporally consistent algorithms.

\begin{table}[h]
\caption{Comparison of main hyperparameters of each algorithm. Inc-TCSR can be deemed a special case of D-TCSR where $\tau=1$.}
\label{tab:comparsion_algs}
\vskip 0.15in
\begin{center}
\begin{sc}
\begin{tabular}{lcccr}
\multicolumn{1}{c}{\bf ALGORITHM}        & \multicolumn{1}{c}{\bf $\ell$ in MLE} & \multicolumn{1}{c}{$\lambda$}  & \multicolumn{1}{c}{$\tau$}       & \multicolumn{1}{c}{\bf BATCH?} \\
\hline \\
SA Init St.      & $0$ &  \---                &   \---       & $\surd$ \\
SA Landm.        & $>0$ &  $1$                &   \---       & $\surd$ \\
TCSR (fitted)    & $>0$ &  $\in[0,1]$         &   \---       & $\times$ \\
Inc-TCSR         & $>0$ &  $\in[0,1]$         &   $1$        & $\surd$ \\
D-TCSR           & $>0$ &  $\in[0,1]$         &   $\in(0,1]$ & $\surd$ \\
\end{tabular}
\end{sc}
\end{center}
\vskip -0.1in
\end{table}


\section{Further experimental details}

\subsection{Metrics for the small-data protocol}
\label{app:ev_metrics}

We use two standard scores for evaluating survival models, namely the Concordance Index (CI) and Integrated Brier Score (IBS).

The CI measures how well the model ranks subjects according to their survival time. 
It is defined as follows:
\begin{equation*}
    \label{eq:CI}
    \mathrm{CI}(\cD) = P\big\{S(t^{(i)}|x^{(i)}_0) > S(t^{(j)}|x^{(j)}_0) : t^{(i)} > t^{(j)}\big\}\,.
\end{equation*}
A CI close to $1$ indicates that the model captures the right order in which the subjects reach the terminal state or censoring time. In contrast, a CI close to $0.5$ suggests the model is not better than a random ranking.

Another popular metric to assess survival models is the IBS, which measures the overall performance of the model across a specific time horizon.
The IBS is calculated by averaging the Brier Score over time.
Given a dataset $\cD$ the Brier Score (BS) is defined as
\begin{equation*}
    \label{eq:BS}
    \begin{aligned}
        \mathrm{BS}(k;\cD) = \sum_{i=1}^n \Bigg[\frac{S(k|x^{(i)}_0)^2}{\mathrm{km}(t^{(i)})} \mathbf{1}\{k\le t^{(i)} \wedge\neg c^{(i)}\}
        + \frac{\big(1 - S(k|x^{(i)}_0)^2\big)}{\mathrm{km}(k)}\mathbf{1}\{k > t^{(i)}\}\Bigg] \,.
    \end{aligned}
\end{equation*}
where $\mathrm{km}$ is the Kaplan–Meier estimator.

The BS is an error score that measures the discrepancy between the estimated survival probability and the true survival outcome. 
Integrating Equation~\eqref{eq:BS} over the time gives the IBS:
\begin{equation*}
    \label{eq:IBS}
    \mathrm{IBS}(\cD) = \frac{1}{H}\sum_{k=0}^{H-1} BS(k;\cD)\,.
\end{equation*}

\subsection{Time-dependent evaluation Metrics}
\label{appendix:timedep_ev_metrics}

Following the landmarking evaluation of Dynamic-DeepHit \citep{lee2019dynamic},
we assess predictions issued at a landmark time $t$ for a prediction window
(horizon) $\Delta$, both measured in discrete time steps. At landmark $t$, only
subjects still at risk ($T > t$) are scored, using the state observed at the
landmark, $x_t$, and follow-up times are re-expressed relative to the landmark,
$\tilde T = T - t$, with $c = 0$ for an observed event and
$c = 1$ for right censoring. For subject $i$, the model's discrete hazards at
that state yield the conditional survival
\[
\hat S_i(\Delta \mid t)
=
\prod_{k=1}^{\Delta}
\bigl[1-h_\theta(k \mid x_t^{(i)})\bigr],
\]
an estimate of
$\Pr(T^{(i)} > t+\Delta \mid T^{(i)} > t, x_t^{(i)})$, and the conditional
risk $\hat F_i(\Delta \mid t) = 1-\hat S_i(\Delta \mid t)$.

Discrimination is measured with the time-dependent concordance index
\begin{equation*}
C(t,\Delta) \;=\;
\frac{\sum_{i \neq j} A_{ij}\, N_i(\Delta)\, \kappa_{ij}}
     {\sum_{i \neq j} A_{ij}\, N_i(\Delta)},
\qquad
A_{ij} = \1[\tilde T^{(i)} < \tilde T^{(j)}],\;
N_i(\Delta) = \1[\tilde T^{(i)} \le \Delta,\, c^{(i)} = 0],
\end{equation*}
where
$\kappa_{ij}
= \1[\hat F_i(\Delta\mid t) > \hat F_j(\Delta\mid t)]
+ \tfrac{1}{2}\,\1[\hat F_i(\Delta\mid t) = \hat F_j(\Delta\mid t)]$.
A pair is comparable when the subject with the earlier time had an observed
event within the window, and the index is the fraction of comparable pairs
ranked correctly by the predicted risk, with ties counted as one half. Since
$C(t,\Delta)$ depends on $\hat F$ only through its ranking, we rank subjects by
$-\log \hat S_i(\Delta \mid t)$, which induces the same order but avoids
numerical underflow when $\hat S_i$ approaches zero at long horizons. Cells
without comparable pairs are undefined and excluded from averages.

Calibration is measured with the inverse-probability-of-censoring-weighted
(IPCW) Brier score \citep{graf1999assessment,gerds2006consistent},
\begin{equation*}
\mathrm{BS}(t,\Delta)
=
\frac{1}{n_t}
\sum_{i:\,T^{(i)} > t}
\left[
\frac{
\1[\tilde T^{(i)} \le \Delta,\, c^{(i)} = 0]
}{
\hat G_t(\tilde T^{(i)})
}
\hat S_i(\Delta \mid t)^2
+
\frac{
\1[\tilde T^{(i)} > \Delta]
}{
\hat G_t(\Delta)
}
\bigl(1-\hat S_i(\Delta \mid t)\bigr)^2
\right],
\end{equation*}
where $n_t$ is the number of subjects at risk at $t$ and $\hat G_t$ is the
Kaplan--Meier estimate of the censoring survival function, fitted on the
training subjects at risk at $t$ with times measured from the landmark.
Subjects censored within the window have unknown status at $t+\Delta$ and
receive zero weight. The weights $1/\hat G_t$ account for this missing outcome
information under the independent-censoring assumption. Simply discarding
subjects censored within the prediction window, or assigning them an outcome at
$t+\Delta$, can bias the resulting score.

Landmarks are fixed per dataset at the quartiles of the observed event or
censoring times, and the horizons at each landmark are the quartiles of the
remaining time $T-t$ among the subjects at risk. This grid is computed once per
dataset and shared by all models and seeds, so that every $(t,\Delta)$ cell can
be averaged across runs.

\subsection{Landmark-based protocol results}
\label{app:fine_grained_results}

In this Section we present the results for the experimental protocol described in Section~\ref{subsec:large_datasets}.
For all models and datasets we present the average metric across 30 seeds with bars representing one standard error.
Results that are significantly better than the other baselines are highlighted with stars. 
Those where D-TCSR significantly beats Inc-TCSR are marked with a circle, and with a dagger in the case they are significantly worse.
We used a paired $t$-test across seeds to compare results.

\begin{figure}[h]

\includegraphics[width=\textwidth]{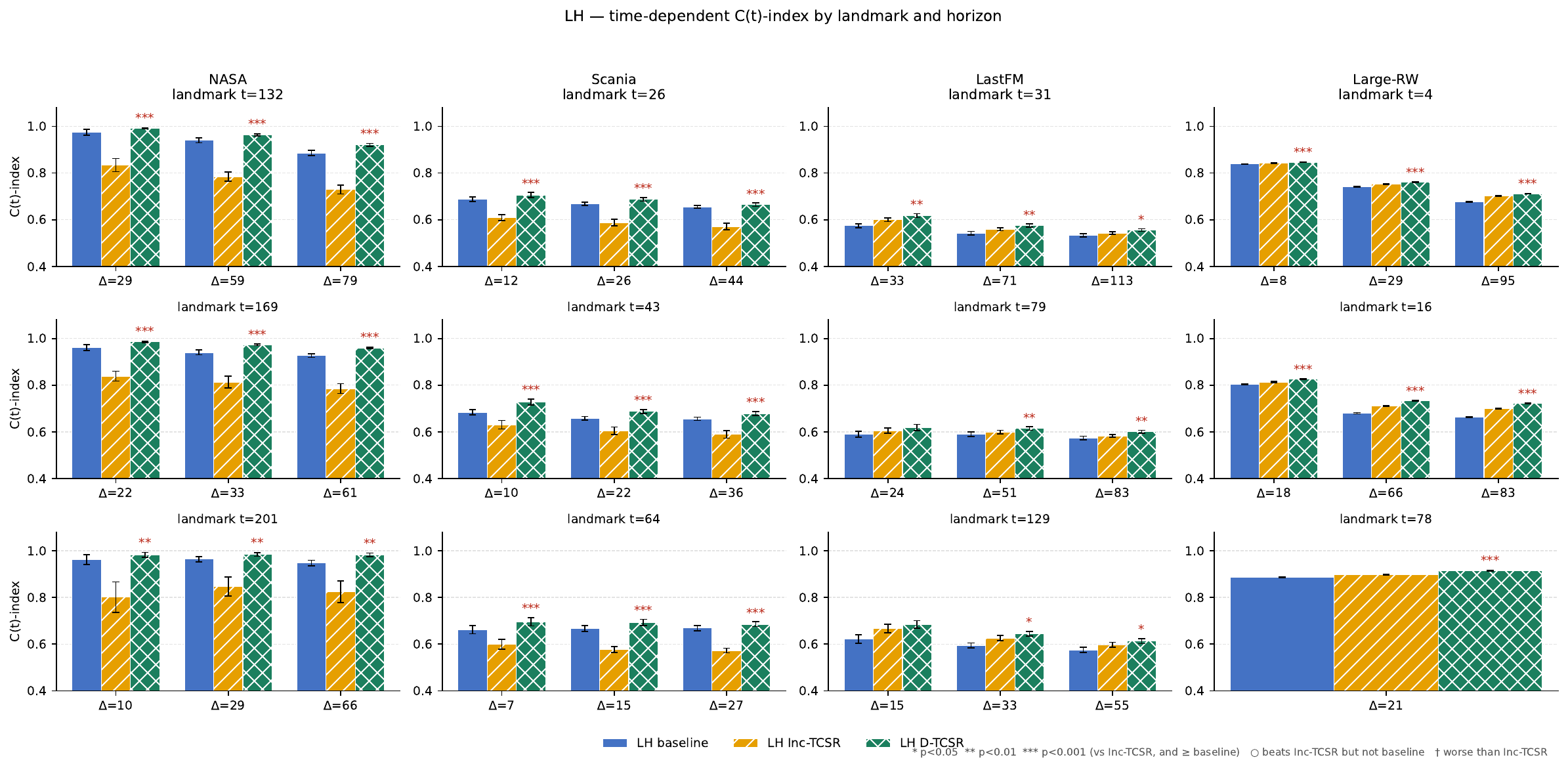}

\caption{Concordance Index (higher is better) of event prediction in big datasets for the LH model for each landmark and predictive horizon. All the results are based on best validation score and reported over 30 random seeds. Results that are significantly better than the baselines based on a paired $t$-test are highlighted with stars.}
\label{fig:lh_ci}
\end{figure}

\begin{figure}[h]

\includegraphics[width=\textwidth]{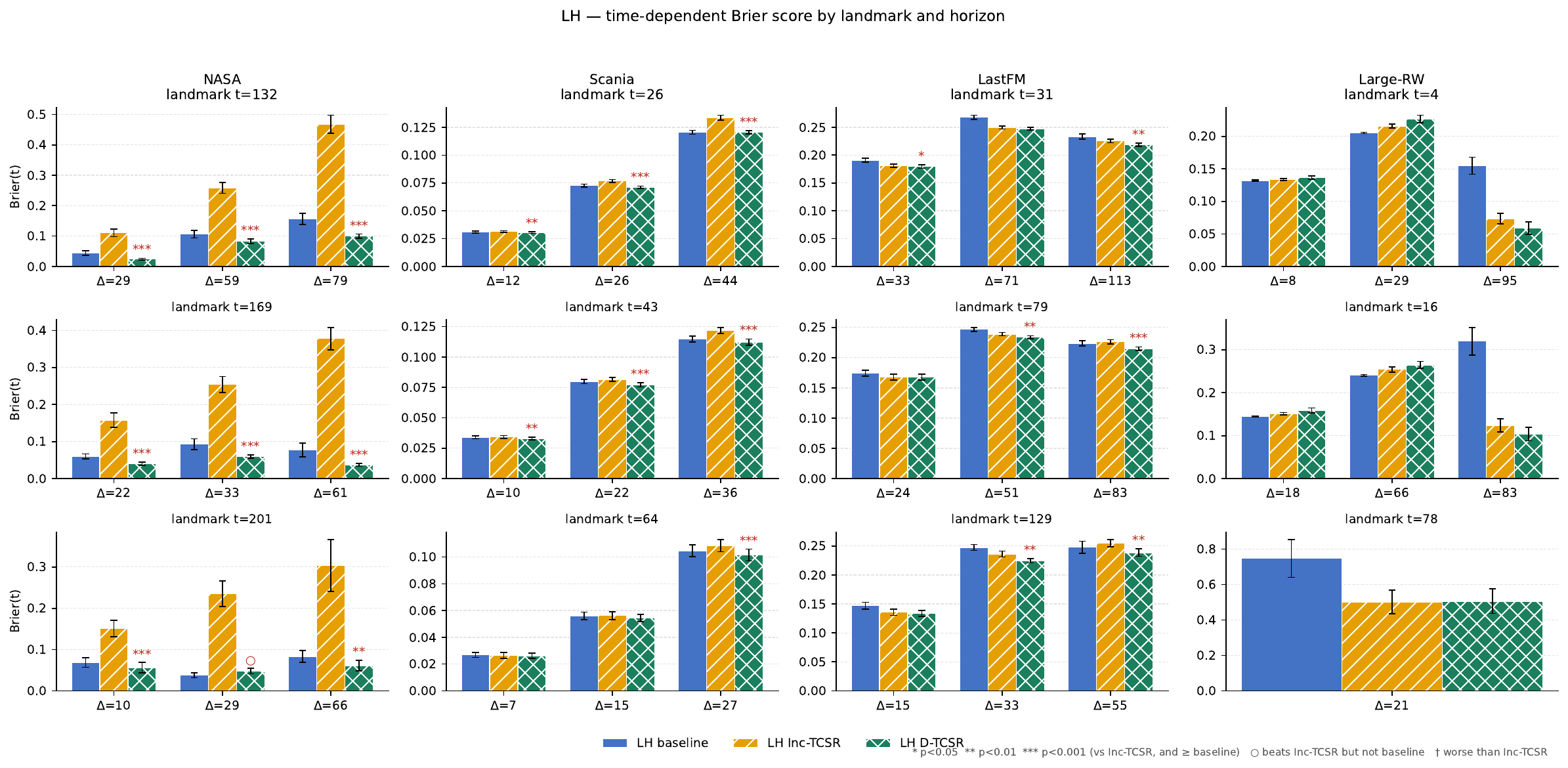}

\caption{Brier score (lower is better) of event prediction in big datasets for the LH model for each landmark and predictive horizon. All the results are based on best validation score and reported over 30 random seeds. Results that are significantly better than the baselines based on a paired $t$-test are highlighted with stars.}
\label{fig:lh_brier}
\end{figure}

Figure~\ref{fig:lh_ci} and Figure~\ref{fig:lh_brier} summarize the results for the CI and BS for each landmark--predictive horizon cell for the LH model.
In terms of CI, D-TCSR systematically outperforms the baselines for every dataset, landmark and predictive horizon, significantly so for most of cells.
As for the BS, the results are more mixed: in NASA the advantage is clear, whereas in LargeRW there are no significant wins for D-TCSR.

\begin{figure}[h]

\includegraphics[width=\textwidth]{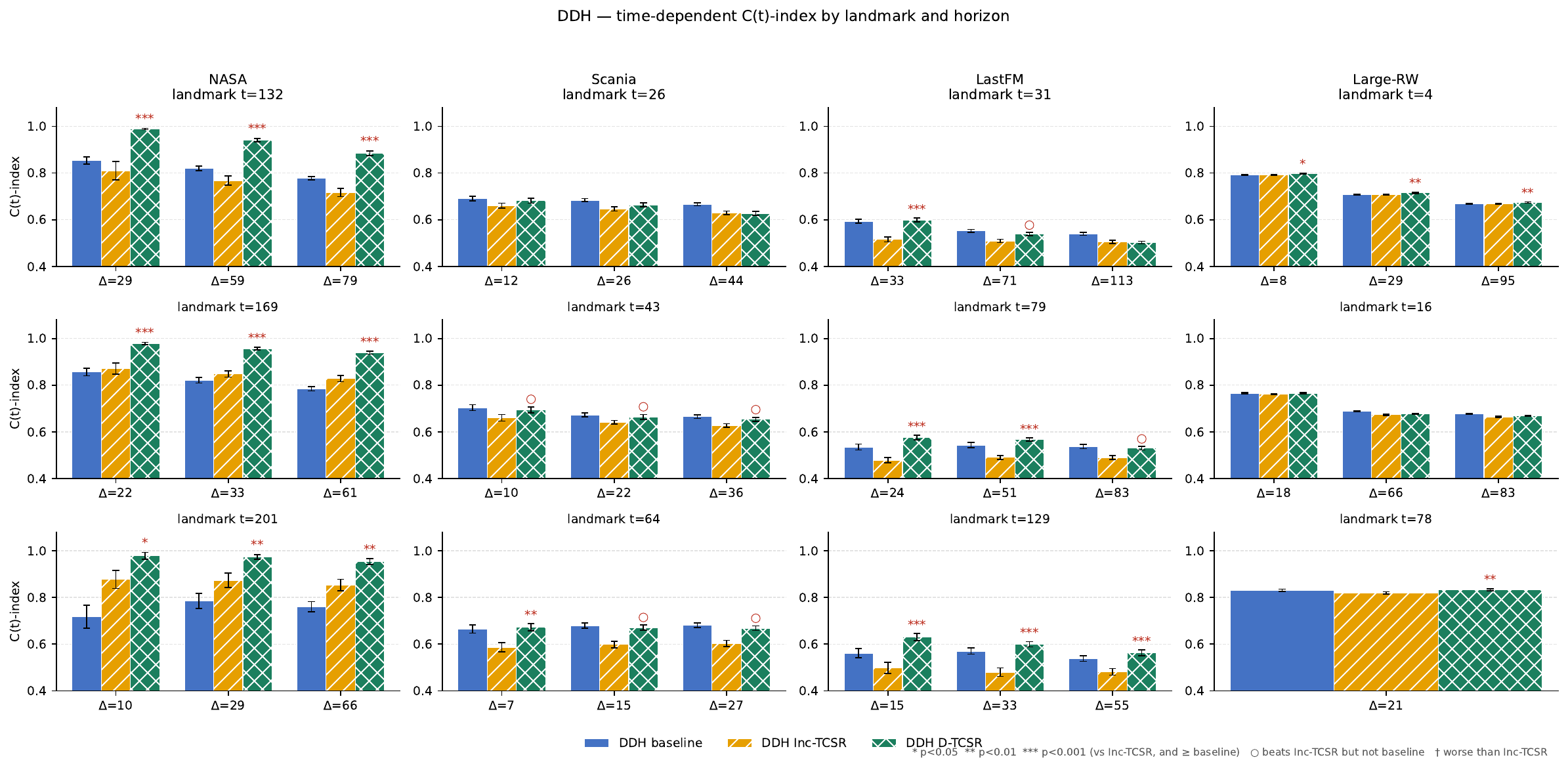}

\caption{Concordance Index (higher is better) of event prediction in big datasets for the DDH model for each landmark and predictive horizon. All the results are based on best validation score and reported over 30 random seeds. Results that are significantly better than the baselines based on a paired $t$-test are highlighted with stars.}
\label{fig:ddh_ci}
\end{figure}

\begin{figure}[h]

\includegraphics[width=\textwidth]{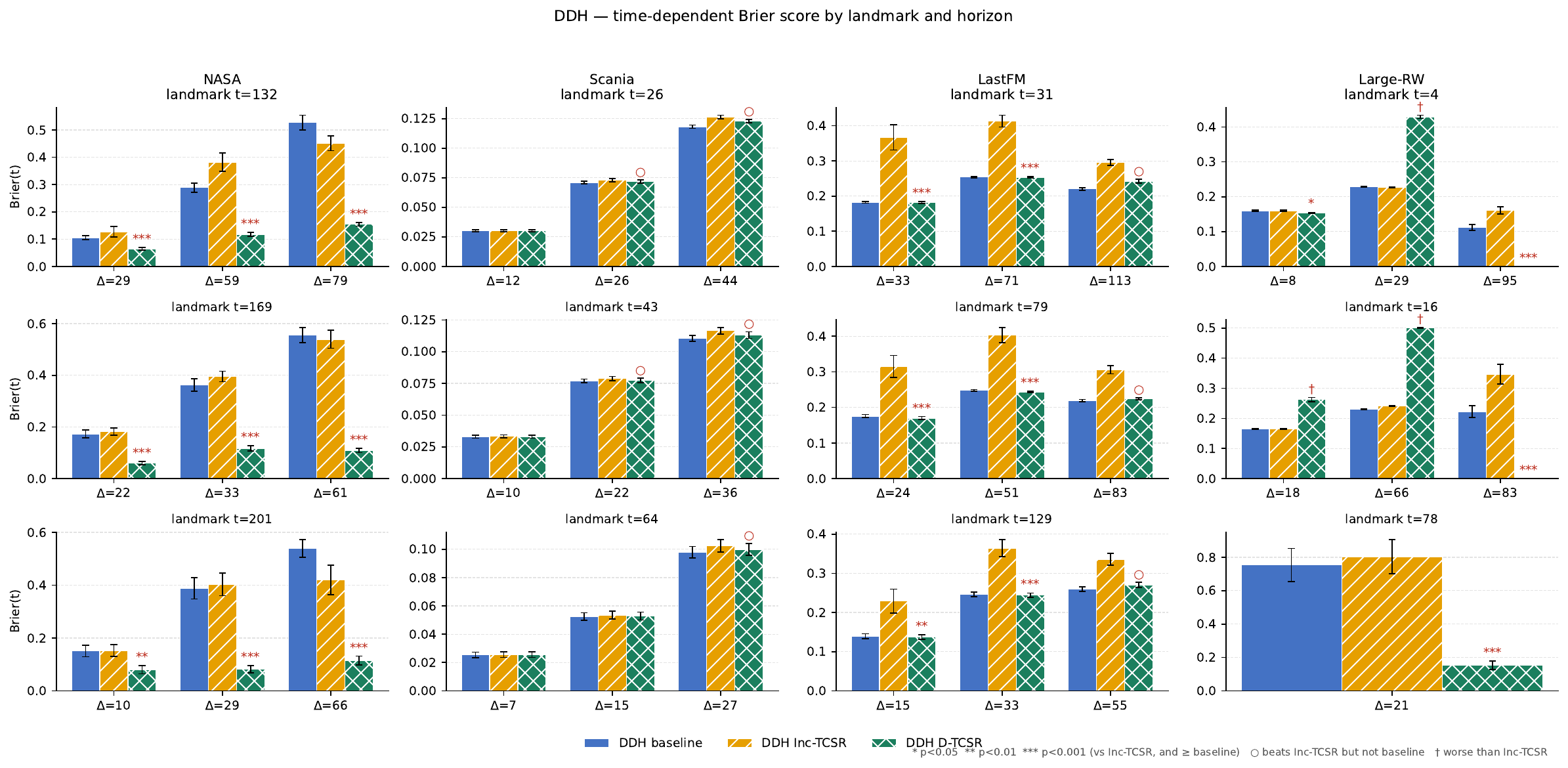}

\caption{Brier score (lower is better) of event prediction in big datasets for the DDH model for each landmark and predictive horizon. All the results are based on best validation score and reported over 30 random seeds. Results that are significantly better than the baselines based on a paired $t$-test are highlighted with stars.}
\label{fig:ddh_brier}
\end{figure}

DDH is a more powerful model that might benefit from temporal consistency to a lesser extent.
However, in NASA and LastFM D-TCSR significantly outperforms all baselines in most cases, and Int-TCSR in some cells, in terms of both CI and BS.
In SCANIA and LargeRW these wins are more nuanced, with significant wins of D-TCSR over Inc-TCSR in cells where temporal consistency does not have a clear advantage over the baseline method.
Results for the DDH model are presented in Figures~\ref{fig:ddh_ci} and \ref{fig:ddh_brier}.

In the overall, these results highlight the benefits of bringing temporal consistency to survival models and of separating the two update timescales through a delayed target network.

\subsection{Architecture details}
\label{appendix:arch_details}

As mentioned in Section~\ref{subsec:large_datasets}, we tune the hyperparameters of both families in two stages.
First we tune the architecture keeping the temporal consistency hyperparameters fixed, and then we fix the architecture and tune $\lambda$ and $\tau$.
In both cases sequences are split at random into 60\%/20\%/20\% training, validation
  and test sets, stratified by censoring status, with a different split for
  each seed. 

For the first stage, each family uses one architecture for all datasets and arms, chosen in a pilot
  on validation data.
For LH, a causal pre-norm transformer encoder \citep{vaswani2017attention} with 2 layers of
  width 64, 4 heads, and dropout set to 0.2.
  For DDH, a Gated Recurrent Unit (GRU) of width 128 with temporal
  attention.
  All arms use AdamW \citep{loshchilov2018decoupled} with learning rate
  $10^{-2}$ and weight decay $10^{-4}$.
  For all datasets except NASA, we use a mini-batch size of 128 sequences. For NASA we reduce this number to 64.
  We train all models for at most 1{,}000 epochs with early stopping with a patience value of 5 steps on the validation
  loss. 
  Table~\ref{tab:arch} presents details of the architecture grid search and winning configurations.

  For tuning the temporal consistency hyperparameters $\lambda$ and $\tau$, we fix the architecture for both models. 
  Inc-TCSR and D-TCSR share the same implementation and tune
  $\lambda \in \{0.1, 0.5, 0.95\}$.
  Inc-TCSR can be deemed a variant of D-TCSR where $\tau$ is fixed as $\tau=1$.
  D-TCSR additionally tunes
  $\tau \in \{0.001, 0.003, 0.01, 0.05, 0.1, 0.25, 0.5, 0.75, 0.9, 0.95, 0.99\}$. 
  One configuration per (method,
  dataset, metric) is selected on the mean validation score over seeds 0--9: the
  validation C-index or IBS for the state-0 protocol, and the validation
  $C(t,\Delta)$ or $\mathrm{BS}(t,\Delta)$, averaged over all landmark
  ($t > 0$) and horizon cells, for the landmark protocol. 
  Each selected
  configuration is trained on 30 seeds, and we report its mean test
  metric ($\pm$ one standard error), with paired $t$-tests
  across seeds.

\begin{table}[t]
  \caption{Selected architectures. Italics: values searched in the pilot.
  Both families use AdamW (learning rate $10^{-2}$, weight decay $10^{-4}$) and
  early stopping on the validation loss.}
  \label{tab:arch}
  \begin{center}
  \begin{tabular}{lll}
   & \multicolumn{1}{c}{\textbf{LH}} & \multicolumn{1}{c}{\textbf{DDH}} \\
  \hline \\
  Encoder        & causal transformer (pre-norm, residual) & GRU + temporal attention \\
  Width          & 64 \;\textit{(of 64, 128)}              & 128 \;\textit{(of 64, 128)} \\
  Depth          & 2 layers \;\textit{(of 2, 4)}           & 1 GRU layer \\
  Attention      & 4 heads, key size 16                    & additive, width 16, over past states \\
  Dropout        & 0.2                                     & -- \\
  Output head    & linear $\to$ $H$ hazard logits          & linear $\to$ $H$ hazard logits \\
  Auxiliary loss & --                                      & ranking (1.0) + covariate prediction (0.1) \\
  \end{tabular}
   \end{center}
\end{table}

\end{document}